\documentclass{llncs}

\usepackage{makeidx}
\usepackage{graphicx}
\usepackage{url}
\usepackage{algpseudocode}
\usepackage[ruled,section]{algorithm}
\usepackage{booktabs}
\usepackage{enumitem}
\usepackage{multirow}
\usepackage{amsmath}

\usepackage{xcolor}

\definecolor{darkgreen}{RGB}{0,127,0}

\begin{document}
\frontmatter          % for the preliminaries
\pagestyle{headings}  % switches on printing of running heads
\addtocmark{Hamiltonian Mechanics} % additional mark in the TOC

\title{Fast ALS-based tensor factorization for context-aware recommendation from implicit feedback}

\author{Bal\'azs Hidasi\inst{1,}\inst{2,}\thanks{B. Hidasi was supported by T\'AMOP-4.2.2.B-10/1--2010-0009.} \and Domonkos Tikk\inst{1}}

\authorrunning{Bal\'azs Hidasi et al.} % abbreviated author list (for running head)

\institute {Gravity R\&D Ltd. \and Budapest University of Technology and Economics
\email{\{balazs.hidasi,domonkos.tikk\}@gravityrd.com}}

\maketitle              % typeset the title of the contribution

\begin{abstract} \small\baselineskip=9pt
Albeit the implicit feedback based recommendation problem---when only the user history is available but there are no ratings---is the most typical setting in real-world applications, it is much less researched than the explicit feedback case. State-of-the-art algorithms that are efficient on the explicit case cannot be straightforwardly transformed to the implicit case if scalability should be maintained. There are few implicit feedback benchmark datasets, therefore new ideas are usually experimented on explicit benchmarks. In this paper, we propose a generic context-aware implicit feedback recommender algorithm, coined iTALS. iTALS applies a fast, ALS-based tensor factorization learning method that scales linearly with the number of non-zero elements in the tensor. The method also allows us to incorporate various contextual information into the model while maintaining its computational efficiency. We present two context-aware implementation variants of iTALS. The first incorporates seasonality and enables to distinguish user behavior in different time intervals. The other views the user history as sequential information and has the ability to recognize usage pattern typical to certain group of items, e.g. to automatically tell apart product types that are typically purchased repetitively or once.
Experiments performed on five implicit datasets (LastFM 1K, Grocery, VoD, and ``implicitized'' Netflix and MovieLens 10M) show that by integrating context-aware information with our factorization framework into the state-of-the-art implicit recommender algorithm the recommendation quality improves significantly. 
\end{abstract}

\section{Introduction}\label{sec:intr}

Recommender systems are information filtering algorithms that help users in information overload to find interesting items (products, content, etc). Users get personalized recommendations that contain typically a few items deemed to be of user's interest. The relevance of an item with respect to a user is predicted by recommender algorithms; items with the highest prediction scores are displayed to the user.

Recommender algorithms are usually sorted into two main approaches: the content based filtering (CBF) and the collaborative filtering (CF). CBF algorithms use user metadata (e.g. demographic data) and item metadata (e.g. author, genre, etc.) and predict user preference using these attributes. In contrast, CF methods do not use metadata, but only data of user--item interactions. Depending on the nature of the interactions, CF algorithms can be further classified into explicit and implicit feedback based methods. In the former case, users provide explicit information on their item preferences, typically in form of user ratings. In the latter case, however, users express their item preferences only implicitly, as they regularly use an online system; typical implicit feedback types are viewing and purchasing. Obviously, implicit feedback data is less reliable as we will detail later. CF algorithms proved to be more accurate than CBF methods, if sufficient preference data is available \cite{PilaRecsys09}.

CF algorithms can be classified into memory-based and model-based ones. Until recently, memory-based solutions were concerned as the state-of-the-art. These are neighbor methods that make use of item or user rating vectors to define similarity, and they calculate recommendations as a weighted average of similar item or user rating vectors. In the last few years, model-based methods gained enhanced popularity, because they were found to be much more accurate in the Netflix Prize, a community contest launched in late 2006 that provided the largest explicit benchmark dataset (100M ratings) \cite{Netflix} for a long time.

Model-based methods build generalized models that intend to capture user preference. The most successful approaches are the latent factor algorithms. These represent each user and item as a feature vector and the rating of user $u$ for item $i$ is predicted as the scalar product of these vectors. Different matrix factorization (MF) methods are applied to compute these vectors, which approximate the partially known rating matrix using alternating least squares (ALS) \cite{BellkorICDM07}, gradient \cite{brismf} and coordinate descent method \cite{Recsys10}, conjugate gradient method \cite{Recsys11}, singular value decomposition \cite{KorenKDD08}, or a probabilistic framework \cite{Salak08}.

Explicit feedback based methods are able to provide accurate recommendations if enough ratings are available. In certain application areas, such as movie rental, travel applications, video streaming, users have motivation to provide ratings to get better service, better recommendations, or award or punish a certain vendor. However, in general, users of an arbitrary online service do not tend to provide ratings on items even if such an option is available, because (1) when purchasing they have no information on their satisfaction (2) they are not motivated to return later to the system to rate. In such cases, user preferences can only be inferred by interpreting user actions (also called \emph{events}). For instance, a recommender system may consider the navigation to a particular product page as an implicit sign of preference for the item shown on that page \cite{RicciRSH11}. The user history specific to items are thus considered as implicit feedback on user taste. Note that the interpretation of implicit feedback data may not necessarily reflect user satisfaction which makes the implicit feedback based preference modeling a difficult task. For instance, a purchased item could be disappointing for the user, so it might not mean a positive feedback. We can neither interpret missing navigational or purchase information as negative feedback, that is, such information is not available.

Despite its practical importance, this harder but more realistic task has been less studied. The proposed solutions for the implicit task are often the algorithms for the explicit problems that had been modified in a way that they can handle the implicit task.

The classical MF methods only consider user-item interaction (ratings or events) when building the model. However, we may have additional information related to items, users or events, which are together termed \emph{contextual information}, or briefly \emph{context}. Context can be, for instance, the time or location of recommendation, social networks of users, or user/item metadata \cite{AdomaviciusRecsys08}. Integrating context can help to improve recommender models. Tensor factorization have been suggested as a generalization of MF for considering contextual information \cite{KaratzogluRecsys10}. However, the existing methods only work for the explicit problem. In this work, we developed a tensor factorization algorithm that can efficiently handle the implicit recommendation task.

The novelty of our work is threefold: (1) we developed a fast tensor factorization method---coined iTALS---that can efficiently factorize huge tensors; (2) we adapted this general tensor factorization to the implicit recommendation task; (3) we present two specific implementations of this general implicit tensor factorization that consider different contextual information. The first variant uses seasonality which was also used in \cite{KaratzogluRecsys10} for the explicit problem. The second algorithm applies sequentiality of user actions and is able to learn association rule like usage patterns. By using these patterns we can tell apart items or item categories having been purchased with different repetitiveness, which improves the accuracy of recommendations. To our best knowledge, iTALS is the first factorization algorithm that uses this type of information.

This paper is organized as follows. Section~\ref{sec:rel} briefly reviews related work on context-aware recommendation algorithms and tensor factorization. In Section~\ref{sec:algo} we introduce our tensor factorization method and its application to the implicit recommendation task. Section~\ref{sec:derived} shows two application examples of our factorization method: (1) we show how seasonality can be included in recommendations and (2) we discuss how a recommendation algorithm can learn repetitiveness patterns from the dataset. Section~\ref{sec:numres} presents the results of our experiments, and Section~\ref{sec:conc} sums up our work and derive the conclusions. 
\subsection{Notation}\label{sec:not}
We will use the following notation in the rest of this paper:
\begin{itemize}[noitemsep,topsep=0pt,parsep=0pt,partopsep=0pt]
    \item $A\circ B\circ\ldots \rightarrow $ The Hadamard (elementwise) product of $A$, $B$, \ldots The operands are of equal size, and the result's size is also the same. The element of the result at $(i,j,k,\ldots)$ is the product of the element of $A$, $B$, \ldots at $(i,j,k,\ldots)$.  This operator has higher precedence than matrix multiplication in our discussion.
    \item $A_{\bullet,i} / A_{i,\bullet} \rightarrow $ The $i^{\rm th}$ column/row of matrix $A$.
    \item $A_{i_1,i_2,\ldots} \rightarrow $ The $(i_1,i_2,\ldots)$ element of tensor/matrix $A$.
    \item $K \rightarrow $ The number of features, the main parameter of factorization.
    \item $D \rightarrow $ The number of dimensions of the tensor.
    \item $T \rightarrow $ A $D$ dimensional tensor that contains only zeroes and ones (preference tensor).
    \item $W \rightarrow $ A tensor with the exact same size as $T$ (weight tensor).
    \item $S_i \rightarrow $ The size of $T$ in the $i^{\rm th}$ dimension ($i=1,\ldots, D$).
    \item $N^+ \rightarrow $ The number of non-zero elements in tensor $T$.
    \item $M^{(i)} \rightarrow $ A $K\times S_i$ sized matrix. Its columns are the feature vectors for the entities in the $i^{\rm th}$ dimension.
\end{itemize} 
\section{Related work}\label{sec:rel}

Context-aware recommender systems \cite{AdomaviciusACMTIS05} emerged as an important research topic in the last years and entire workshops are devoted to this topic on major conferences (CARS series started in 2009 \cite{CARS2009}, CAMRA in 2010 \cite{CAMRA2010}). The application fields of context-aware recommenders include among other movie \cite{BogersCARS10} and music recommendation \cite{BaltrunasCARS09}, point-of-interest recommendation (POI) \cite{BaderCRR11}, citation recommendation \cite{HeWWW10}. Context-aware recommender approaches can be classified into three main groups: pre-filtering, post-filtering and contextual modeling \cite{AdomaviciusRecsys08}. Baltrunas and Amatriain \cite{BaltrunasCARS09} proposed a pre-filtering approach by partitioned user profiles into \emph{micro-profiles} based on the time split of user event falls, and experimented with different time partitioning. Post-filtering ignores the contextual data at recommendation generation, but disregards irrelevant items (in a given context) or adjust recommendation score (according to the context) when the recommendation list is prepared; see a comparison in \cite{PannielloRecsys09}. The tensor factorization based solutions, including our proposed approach, falls into the contextual modeling category.

Tensor factorization incorporates contextual information into the recommendation model. Let us have a set of items, users and ratings (or events) and assume that additional context of the ratings is available (e.g.\ time of the rating). Having $C$ different contexts, the rating data can be cast into a $C+2$ dimensional tensor. The first dimension corresponds to users, the second to items and the subsequent $C$ dimensions $[3,\ldots,C+2]$ are devoted to contexts. We want to decompose this tensor into lower rank matrices and/or tensors in a way that the reconstruction the original tensor from its decomposition approximates well the original tensor. Approximation accuracy is calculated at the known positions of the tensor using RMSE as error measure. In \cite{KaratzogluRecsys10}, a sparse HOSVD \cite{hosvd} method is presented that decomposes a $D$ dimensional sparse tensor into $D$ matrices and a $D$ dimensional tensor. If the size of the original tensor is $S_1\times S_2\times \cdots\times S_{D}$ and the number of features is $K$ then the size of the matrices are $S_1\times K$, $S_2\times K$, \dots, $S_{D}\times K$ and the size of the tensor is $K\times K\times\cdots\times K$. The authors use gradient descent on the known ratings to find the decomposition, and by doing so, the complexity of one iteration of their algorithm scales \emph{linearly} with the number of non-missing values in the original tensor (number of rating) and \emph{cubically} with the number of features ($K$). This is much less than the cost of the dense HOSVD, which is $O(K\cdot (S_1+\cdots+S_{D})^{D})$. A further improvement was proposed by Rendle \emph{et al} \cite{RendleSIGIR11}, where the computational complexity was reduced so that their method scales linearly \emph{both} with the number of explicit ratings and with the number of features. However, if the original tensor is large and dense like for the implicit recommendation task then neither method scales well.

\section{ALS based fast tensor factorization}\label{sec:algo}

In this section we present iTALS, a general ALS-based tensor factorization algorithm that scales linearly with the non-zero element of a dense tensor (when appropriate weighting is used) and cubically with the number of features. This property makes our algorithm suitable to handle the context-aware implicit recommendation problem.

Let $T$ be a tensor of zeroes and ones and let $W$ contain weights to each element of $T$. $T_{u,i,c_1,\cdots,c_C}$ is $1$ if user $u$ has (at least one) event on item $i$ while the context-state of $j^{\rm{th}}$ context dimension was $c_j$, thus the proportion of ones in the tensor is very low. An element of $W$ is $1$ if the corresponding element in $T$ is $0$ and greater than $1$ otherwise. Instead of using the form of the common HOSVD decomposition ($D$ matrices and a $D$ dimensional tensor) we decompose the original $T$ tensor into $D$ matrices. The size of the matrices are $K\times S_1, K\times S_2, \ldots, K\times S_{D}$. The prediction for a given cell in $T$ is the elementwise product of columns from $M^{(i)}$ low rank matrices. Equation \ref{eq:reconstruct} describes the model.

\begin{equation}\label{eq:reconstruct}
\hat{T}_{i_1,i_2,\ldots,i_{D}}=1^TM^{(1)}_{\bullet,i_1}\circ M^{(2)}_{\bullet,i_2}\circ\cdots\circ M^{(D)}_{\bullet,i_{D}}
\end{equation}

We want to minimize the loss function of equation \ref{eq:loss}:

\begin{equation}\label{eq:loss}
L(M^{(1)},\ldots,M^{(D)})=\sum_{i_1=1,\ldots,i_{D}=1}^{S_1,\ldots,S_{D}}W_{i_1,\ldots,i_{D}}\left(T_{i_1,\ldots,i_{D}}-\hat{T}_{i_1,\ldots,i_{D}}\right)^2
\end{equation}

If all but one $M^{(i)}$ is fixed, $L$ is convex in the non-fixed variables. We use this method to minimize the loss function. $L$ reaches its minimum (in $M^{(i)}$) where its derivate with respect to $M^{(i)}$ is zero. Since the derivate of $L$ is linear in $M^{(i)}$ the columns of the matrix can be computed separately. For the $(i_1)^{\rm{th}}$ column of $M^{(1)}$:

\begin{equation}\label{eq:diff}
\begin{gathered}
0=\frac{\partial L}{\partial M^{(1)}_{\bullet,i_1}} =
-2\underbrace{\sum_{i_2=1,\ldots,i_{D}=1}^{S_2,\ldots,S_{D}}W_{i_2,
\ldots,i_{D}}T_{i_1,\ldots,i_{D}}\left(M^{(2)}_{\bullet,i_2}\circ\cdots
\circ M^{(D)}_{\bullet,i_{D}}\right)}_{\mathcal{O}} +\\
2\underbrace{\sum_{i_2=1,\ldots,i_{D}=1}^{S_2,\ldots,S_{D}}W_{i_2,
\ldots,i_{D}}\left(M^{(2)}_{\bullet,i_2}\circ\cdots\circ
M^{(D)}_{\bullet,i_{D}}\right)\left(M^{(2)}_{\bullet,i_2}\circ\cdots
\circ
M^{(D)}_{\bullet,i_{D}}\right)^TM^{(1)}_{\bullet,i_1}}_{\mathcal{I}}
\end{gathered}
\end{equation}

It takes $O(DKN^+_{i_1})$ time to compute $\mathcal{O}$ in equation \ref{eq:diff}, because only $N^+_{i_1}$ cells of $T$ for $i_1$ in the first dimension contain ones, the others are zeroes. For every column it yields a complexity of $O(DKN^+)$. The naive computation of $\mathcal{I}$ however is very expensive computationally: $O(K\prod_{i=2}^D{S_i})$. Therefore we transform $\mathcal{I}$ by using $W_{i_2,\ldots,i_{D}}=W'_{i_2,\ldots,i_{D}}+1$ and get:

\begin{equation}\label{eq:difftr}
\begin{gathered}
\mathcal{I}=\sum_{i_2=1,\ldots,i_{D}=1}^{S_2,\ldots,S_{D}}W'_{i_2,\ldots,i_{D}}\left(M^{(2)}_{\bullet,i_2}\circ\cdots\circ M^{(D)}_{\bullet,i_{D}}\right)\left(M^{(2)}_{\bullet,i_2}\circ\cdots\circ M^{(D)}_{\bullet,i_{D}}\right)^TM^{(1)}_{\bullet,i_1}+\\
+\underbrace{\sum_{i_2=1,\ldots,i_{D}=1}^{S_2,\ldots S_{D}}\left(M^{(2)}_{\bullet,i_2}\circ\cdots\circ M^{(D)}_{\bullet,i_{D}}\right)\left(M^{(2)}_{\bullet,i_2}\circ\cdots\circ M^{(D)}_{\bullet,i_{D}}\right)^T}_{\mathcal{J}}M^{(1)}_{\bullet,i_1}
\end{gathered}
\end{equation}

The first part in equation \ref{eq:difftr} can be calculated in $O(K^2N^+_{i_1})$ as $W'_{i_2,\ldots,i_{D}}=(W_{i_2,\ldots,i_{D}}-1)$ and the weights for the zero elements of $T$ are ones.  This step is the generalization of the Hu \emph{et.\ al}'s adaptation of ALS to the implicit problem \cite{HuICDM08}. The total complexity of calculating all columns of the matrix is $O(K^2N^+)$. $\mathcal{J}$ is the same for all columns of $M^{(1)}$ (independent of $i_1$) and thus can be precomputed. However the cost of directly computing $\mathcal{J}$ remains $O(K\prod_{i=2}^D{S_i})$. Observe the following:

\begin{equation}\label{eq:righttr}
\begin{aligned}
\mathcal{J}_{j,k}=&\left(\sum_{i_2=1,\ldots,i_{D}=1}^{S_2,\ldots,S_{D}}\left(M^{(2)}_{\bullet,i_2}\circ\cdots\circ M^{(D)}_{\bullet,i_{D}}\right)\left(M^{(2)}_{\bullet,i_2}\circ\cdots\circ M^{(D)}_{\bullet,i_{D}}\right)^T\right)_{j,k}=\\
=&\sum_{i_2=1,\ldots,i_{D}=1}^{S_2,\ldots,S_{D}}\left(M^{(2)}_{j,i_2}\cdot\ldots\cdot M^{(D)}_{j,i_{D}}\right)\left(M^{(2)}_{k,i_2}\cdot\ldots\cdot M^{(D)}_{k,i_{D}}\right)=\\
=&\left(\sum_{i_2=1}^{S_2}{M^{(2)}_{j,i_2}M^{(2)}_{k,i_2}}\right)\cdot\ldots\cdot\left(\sum_{i_D=1}^{S_D}{M^{(D)}_{j,i_D}M^{(D)}_{k,i_D}}\right)
\end{aligned}
\end{equation}

Using equation \ref{eq:righttr} we can transform the second part from equation \ref{eq:difftr} into the following form:

\begin{equation}\label{eq:finaltr}
\begin{gathered}
\mathcal{J}=\sum_{i_2=1,\ldots,i_{D}=1}^{S_2,\ldots,S_{D}}\left(M^{(2)}_{\bullet,i_2}\circ\cdots\circ M^{(D)}_{\bullet,i_{D}}\right)\left(M^{(2)}_{\bullet,i_2}\circ\cdots\circ M^{(D)}_{\bullet,i_{D}}\right)^T=\\
=\underbrace{\left(\sum_{i_2=1}^{S_2}{M^{(2)}_{\bullet,i_2}\left(M^{(2)}_{\bullet,i_2}\right)^T}\right)}_{\mathcal{M}^{(2)}}\circ\cdots\circ\underbrace{\left(\sum_{i_D=1}^{S_D}{M^{(D)}_{\bullet,i_D}\left(M^{(D)}_{\bullet,i_D}\right)^T}\right)}_{\mathcal{M}^{(D)}}
\end{gathered}
\end{equation}

The members of equation \ref{eq:finaltr} can be computed in $O(S_iK^2)$ time. From the $\mathcal{M}^{(i)}$ matrices the expression can be calculated in $O(K^2D)$ time. Note that the $\mathcal{M}^{(i)}$ is needed for computing all but the $i^{\rm{th}}$ matrix but only changes if $M^{(i)}$ changed. Therefore we count the cost of computing $\mathcal{M}^{(i)}$ to the cost of recomputing $M^{(i)}$. To get the desired column of the matrix we need to invert a $K\times K$ sized matrix per column (see equation \ref{eq:diff}). That requires $O(K^3S_1)$ time for all columns of $M^{(1)}$. The columns of the other matrices can be calculated similarly.

\begin{algorithm}[!h]
    \caption{Fast ALS-based tensor factorization for implicit feedback recommendations}\label{alg:iTALS}
    \textbf{Input:} {$T$: a $D$ dimensional $S_1 \times\cdots \times S_{D}$ sized tensor of zeroes and ones; $W$: a $D$ dimensional $S_1 \times\cdots \times S_{D}$ sized tensor containing the weights; $K$: number of features; $E$: number of epochs} \newline
    \textbf{Output:} {$\{M^{(i)}\}_{i=1,\ldots, D}$} $K\times S_i$ sized low rank matrices \newline
    \textbf{procedure} \Call{iTALS}{$T$, $W$, $K$, $E$}
	\begin{algorithmic}[1]
		\For{$i=1,\ldots,D$}
            \State $M^{(i)} \leftarrow $ Random $K\times S_i$ sized matrix
            \State $\mathcal{M}^{(i)} \leftarrow M^{(i)}(M^{(i)})^T$ \label{step:precomp}
        \EndFor
        \For{$e=1,\ldots,E$}
            \For{$i=1,\ldots,D$}
                \State $C^{(i)} \leftarrow \mathcal{M}^{(\ell_1)}\circ\cdots\circ \mathcal{M}^{(\ell_{D-1})}, (i\not\in\{\ell_1,\ldots,\ell_{D-1}\})$ \label{step:fast}
                \State $T^{(i)} \leftarrow$ \Call{UnfoldTensor}{$T$,$i$}
                \For{$j_i=1,\ldots,S_i$}
                    \State $C^{(i)}_{j_i} \leftarrow C^{(i)}$
                    \State $O^{(i)}_{j_i} \leftarrow 0$
                    \ForAll{$t: \{t\in T^{(i)}_{j_i}, t\neq0\}$} \label{step:col_begin}
                        \State $\{j_\ell|\ell\neq i\} \leftarrow$ Indices of $t$ in $T$
                        \State $W_t \leftarrow$ \Call{GetWeight}{$W$,$t$}
                        \State $v \leftarrow M^{(\ell_1)}\circ\cdots\circ M^{(\ell_{D-1})}, (i\not\in\{\ell_1,\ldots,\ell_{D-1}\})$
                        \State $C^{(i)}_{j_i} \leftarrow C^{(i)}_{j_i} + vW_tv^T$ and $O^{(i)}_{j_i} \leftarrow O^{(i)}_{j_i} + W_tv$
                    \EndFor \label{step:col_end}
                    \State $M^{(i)}_{\bullet,j_i} \leftarrow (C^{(i)}_{j_i}+\lambda I)^{-1}O^{(i)}_{j_i}$ \label{step:reg}
                \EndFor
                \State $\mathcal{M}^{(i)} \leftarrow M^{(i)}(M^{(i)})^T$ \label{step:recomp}
            \EndFor
        \EndFor
        \State \textbf{return} $\{M^{(i)}\}_{i=1\dots D}$
\end{algorithmic}
\textbf{end procedure}
\end{algorithm}

The total cost of computing $M^{(i)}$ is $O(K^3S_i+K^2N^++KDN^+)$ that can be simplified to $O(K^3S_i+K^2N^+)$ using that usually $D\ll K$. Therefore the cost of computing each matrix once is $O\left(K^3\sum_{i=1}^{D}{S_i}+K^2N^+\right)$. Thus the cost of an epoch is linear in the number of the non-zero examples and cubical in the number of features. The cost is also linear in the number of dimensions of the tensor and the sum of the length of the tensors in each dimension. We will also show in Section~\ref{sec:runtimes} that the $O(K^2)$ part is dominant when dealing with practical problems. The complexity of the gradient descent method for implicit feedback is $O(K\prod_{i=1}^{D}{S_i})$ that is linear in the number of features but the $\prod_{i=1}^{D}{S_i}$ part makes impossible to run it on real life datasets. Sampling can be applied to reduce that cost but it is not trivial how to sample in the implicit feedback case.

The pseudocode of the suggested iTALS (Tensor factorization using ALS for implicit recommendation problem) is given in Algorithm~\ref{alg:iTALS}. There we use two simple functions. $\Call{UnfoldTensor}{T,i}$ unfolds tensor $T$ by its $i^{\rm th}$ dimension. This step is used for the sake of clarity, but with proper indexing we would not need to actually unfold the tensor. $\Call{GetWeight}{W,t}$ gets the weight from the weight tensor $W$ for the $t$ element of tensor $T$ and creates a diagonal matrix from it. The size of $W_t$ is $K\times K$ and it contains the weight for $t$ in its main diagonal and $0$ elsewhere. The pseudocode follows the deduction above. In line \ref{step:precomp} we precompute $\mathcal{M}^{(i)}$. We create the column independent part from equation \ref{eq:difftr} in line \ref{step:fast}. We add the column dependent parts to each side of equation \ref{eq:diff} in lines \ref{step:col_begin}--\ref{step:col_end} and compute the desired column in line \ref{step:reg}. In this step we use regularization to avoid numerical instability and overfitting of the model. After each column of $M^{(i)}$ is computed $\mathcal{M}^{(i)}$ is recomputed in line \ref{step:recomp}. 
\section{Context-aware iTALS algorithm}\label{sec:derived}
In this section we derive two specific algorithms from the generic iTALS method presented in Section~\ref{sec:algo}. The first method uses seasonality as context, the second considers the user history as sequential data, and learns meta-rules about sequentiality and repetitiveness.

\subsection{Seasonality}\label{sec:season}
Many application areas of recommender systems exhibit the seasonality effect, therefore seasonal data is an obvious choice as context \cite{LiuCAMRA10}. Strong periodicity can be observed in most of the human activities: as people have regular daily routines, they also follow similar patterns in TV watching at different time of a day, they do their summer/winter vacation around the same time in each year. Taking the TV watching example, it is probable that horror movies are typically watched at night and animation is watched in the afternoon or weekend mornings. Seasonality can be also observed in grocery shopping or in hotel reservation data.

In order to consider seasonality, first we have to define the length of season. During a season we do not expect repetitions in the aggregated behavior of users, but we expect that at the same time offset in different seasons, the aggregated behavior of the users will be similar. The length of the season depends on the data. For example it is reasonable to set the season length to be 1 day for VoD consumption, however, this is not an appropriate choice for shopping data, where 1 week or 1 month is more justifiable. Having the length of the season determined, we need to create \emph{time bands} (bins) in the seasons. These time bands are the possible context-states. Time bands specify the time resolution of a season, which is also data dependent. We can create time bands with equal or different length. For example, every day of a week are time bands of equal length, but 'morning', 'around noon', 'afternoon', 'evening', 'late evening', 'night' could be time bands of a day with different length. Obviously, these two steps require some a-priori knowledge about the data or the recommendation problem, but iTALS is not too sensitive to minor deviations related to the length and the resolution of the season.

In the next step, events are assigned to time bands according to their time stamp. Thus, we can create the (user, item, time band) tensor. We factorize this tensor using the iTALS algorithm and we get feature vectors for each user, for each item and for each time band. When a recommendation is requested for user $u$ at time $t$, first the time band of $t$ is determined and then the preference value for each item using the feature vector of user $u$ and the feature vector of time band $tb_t$ is calculated.

\subsection{Sequentiality}\label{sec:meta}
Recommendation algorithms often recommend items from categories that the user likes. For example if the user often watches horror movies then the algorithm will recommend her horror movies. This phenomenon is even stronger if time decay is applied and so recent events have greater weights. Pushing newer events can increase accuracy, because similar items will be recommended. This functioning can be beneficial in some application fields, like VoD recommendation, but will fail in such cases where repetitiveness in user behavior with respect to items can not be observed. A typical example for that is related to household appliance products: if a user buys a TV set and then she gets further TV sets recommended, she will not probably purchase another one. In such a case, complementary or related goods are more appropriate to recommend, DVD players or external TV-tuners for example. On the other hand, the purchase of a DVD movie does not exclude at all the purchase of another one. Whether recommendation of similar items is reasonable, depends on the nature of the item and behavior of the user. Next, we propose an approach to integrate the repetitiveness of purchase patterns into the latent factor model.

Using association rules is a possible approach to specify item purchase patterns. Association rules \cite{Agrawal93} are often used to determine which products are bought frequently together and it was reported that in certain cases association rule based recommendations yield the best performance \cite{JamesRecsys10}. In our setting, we can extract purchase patterns from the data using association rule mining on the subsequent user events within a given time window.
There are two possibilities: we can generate category--category rules, or category--item rule, thus having usage patterns:
\begin{itemize}
\item if a user bought an item from category $A$ then she will buy an item from category $B$ next time, or
\item if a user bought an item from category $A$ then she will buy an item $X$ next time.
\end{itemize}

We face, however, with the following problems, when attempting to use such patterns in recommendations: (1) the parameter selection (minimum support, minimum confidence and minimum lift) influences largely the performance, their optimization may be slow; (2) rules with negated consequents (e.g. bought from $A$ will not buy from $B$) are not found at all; (3) with category--category rules one should devise further weighting/filtering to promote/demote the items in the pushed category; (4) the category--item rules are too specific therefore either one gets too many rules or the rules will overfit.

We show how repetitiveness related usage patterns can be efficiently integrated into recommendation model using the the iTALS algorithm. Let us now consider the \emph{category of last purchased item} as the context for the next recommendation. The tensor has again three dimensions: users, items and item categories. The $(i,u,c)$ element of the tensor means that user $u$ bought item $i$ and the user's latest purchase (before buying $i$) was an item from category $c$. Using the examples above: the user bought a given DVD player after the purchase of a TV set. After factorizing this tensor we get feature vectors for the item categories as well. These vectors act as weights in the feature space that reweight the user--item relations. For example, assuming that the first item feature means ``having large screen'' then the first feature of the TV category would be low as such items are demoted. If the second item feature means ``item can play discs'' then the second feature of the TV category would be high as these items are promoted.

The advantage of this method is that it learns the usage patterns from the data globally by producing feature vectors that reweight the user--item relations. One gets simple but general usage patterns using the proposed solution that integrates seamlessly into the common factorization framework: no post-processing is required to define promotional/demotional weights/filters.

We can generalize the concept described above to take into account several recent purchases. We could create a $C+2$ dimensional tensor, where the $[3,\ldots,C+2]$ dimensions would represent the item categories of the last $C$ purchases, but the resulting tensor would be very sparse as we increase $C$. Instead we remain at a three dimensional tensor but we set simultaneously $C$ item categories to $1$ for each user--item pair. We may also decrease the weights in $W$ for those additional $C-1$ cells as they belong to older purchases. Thus we may control the effect of previous purchases based on their recency. When recommending, we have to compute the (weighted) average of the feature vectors of the corresponding categories and use that vector as the context feature vector. 
\section{Experiments}\label{sec:numres}
We used five databases to validate our algorithms. Three of them contain genuine implicit feedback data (LastFM 1K and 2 proprietary), while the other two are implicit variants of explicit feedback data. The \textit{LastFM 1K} \cite{lastfm1k} dataset contains listening habits of $\sim$1\,000 users on songs of $\sim$170\,000 artists (artists are considered items). The training set contains all events until 28/04/2009. The test set contains the events of the next day following the training period.
In \textit{VoD consumption dataset}, with 8 weeks of training data we tested on the data of the next day. Thus, all test events occurred after the last train event. The training set contains 22.5 million events and 17\,000 items. The online \emph{grocery} dataset contains only purchase events. We used a few years' data for training and one month for testing. The training set contains 6.24 million events and 14\,000 items. The two explicit feedback datasets are the Netflix \cite{Netflix} and the MovieLens 10M \cite{Movielens}. We kept the five star ratings for the former and ratings of 4.5 and above for the latter and used them as positive implicit feedback. For train-test splits we used the splitting dates 15/12/2005 and 01/12/2008, respectively.

\begin{table}[!t]
\centering
\caption{Recall@20 for all datasets and algorithms using factorization with 20 and 40 features; in each row, the best and second best results are highlighted by bold and slanted typesetting, respectively}
\medskip
{\footnotesize
\begin{tabular}{l cccccc}
\toprule
\multirow{2}{*}{\textbf{Dataset}} & \multirow{2}{*}{\textbf{iALS}}& \textbf{iCA baseline}& \textbf{iTALS} & \textbf{iTALS} & \textbf{iTALS} & \textbf{iTALS} \\
&& \textbf{time bands}& \textbf{time bands}& \textbf{seq.}& \textbf{seq. (2)}& \textbf{seq. (5)}\\
\midrule
VOD (20)&0.0632&\textsl{0.0847}&\textbf{0.1125}&0.0689&0.0678&0.0666\\
VOD (40)&0.0753&\textsl{0.0910}&\textbf{0.1240}&0.0855&0.0930&0.0883\\
\midrule
Grocery (20)&0.0656&0.0803&0.1032&\textbf{0.1261}&\textsl{0.1223}&0.1153\\
Grocery (40)&0.0707&0.0872&0.1081&\textsl{0.1340}&\textbf{0.1351}&0.1189\\
\midrule
LastFM 1K (20)&0.0157&0.0249&0.0352&\textsl{0.0747}&\textbf{0.0793}&0.0733\\
LastFM 1K (40)&0.0333&0.0351&0.0418&0.0785&\textbf{0.0851}&\textsl{0.0800}\\
\midrule
Netflix (20)&0.0540&\textsl{0.0593}&\textbf{0.0724}&0.0512&0.0534&0.0537\\
Netflix (40)&0.0552&\textsl{0.0561}&\textbf{0.0671}&0.0503&0.0527&0.0538\\
\midrule
MovieLens (20)&0.0494&\textsl{0.0553}&\textbf{0.0896}&0.0406&0.0450&0.0457\\
MovieLens (40)&0.0535&0.0494&\textbf{0.0937}&0.0361&0.0480&\textsl{0.0498}\\
\bottomrule
\end{tabular}}
\vspace*{-5pt}
\label{tab:recall}
\end{table}

We determined the seasonality for each dataset, that is, the periodicity patterns observed in the data. As for the VoD data, we defined a day as the season and defined custom time intervals as time bands ('morning', 'around noon', 'afternoon', 'evening', 'late evening', 'night' and 'dawn'), because people watch and channels broadcast different programs at different time of the day. For LastFM 1K and MovieLens we also used a day as the season and time bands of 30 minutes. For the Grocery data we defined a week as the season and the days of the week as the time bands. The argument here is that people tend to follow different shopping behavior on weekdays and weekends. For the Netflix data only the day of the rating is available, so we decided to define a week as the season and the days of the week as time bands.

In our next experiment, we used item category with Grocery and Netflix datasets, genre with VoD and MovieLens and artists for LastFM as the category of the item for the meta-rule learning algorithm. We experimented with using the last 1, 2, 5 events prior to the current event of the users.

We compared the two iTALS variants to the basic iALS as well as to a context-aware baseline for implicit feedback data. This method, referred as \textit{implicit CA (iCA) baseline}, is the composite of several iALS models. For each context state we train a model using only the events with the appropriate context, e.g., with the VoD we train 7 models for the 7 time bands. The context of the recommendation request (e.g. time of day) selects the model for the prediction. This baseline treats context-states independently. Due to its long running time we used iCA only with seasonality, as \#(time bands) $\ll$ \#(preceding item categories).

Every algorithm has three common parameters: the number of features, the number of epochs and the regularization parameter. We set the number of features to 20 and 40 commonly used in literature \cite{PilaRecsys09,KorenKDD08}. The number of epochs was set to 10 as the ranked list of items hardly changes after 10 epochs. The regularization was proportional to the support of the given item/user/context. We did not use any other heuristics like time decay to focus on the pure performance of the algorithms. The weights in $W$ were proportional to the number of events belonging to the given cell of the tensor.

We measured recall and precision on the $N=1,\ldots,50$ interval. We consider items relevant to a user if the user has at least one event for that item in the test set. Recall@$N$ is the ratio of relevant items on the ranked topN recommendations for the user relative to the number of the user's events in the test set. Precision@$N$ is the ratio of the number of returned relevant items (for each user) and the number of total returned items. Greater values mean better performance.

\begin{figure}[!t]
\centering
\includegraphics[width=4.9in]{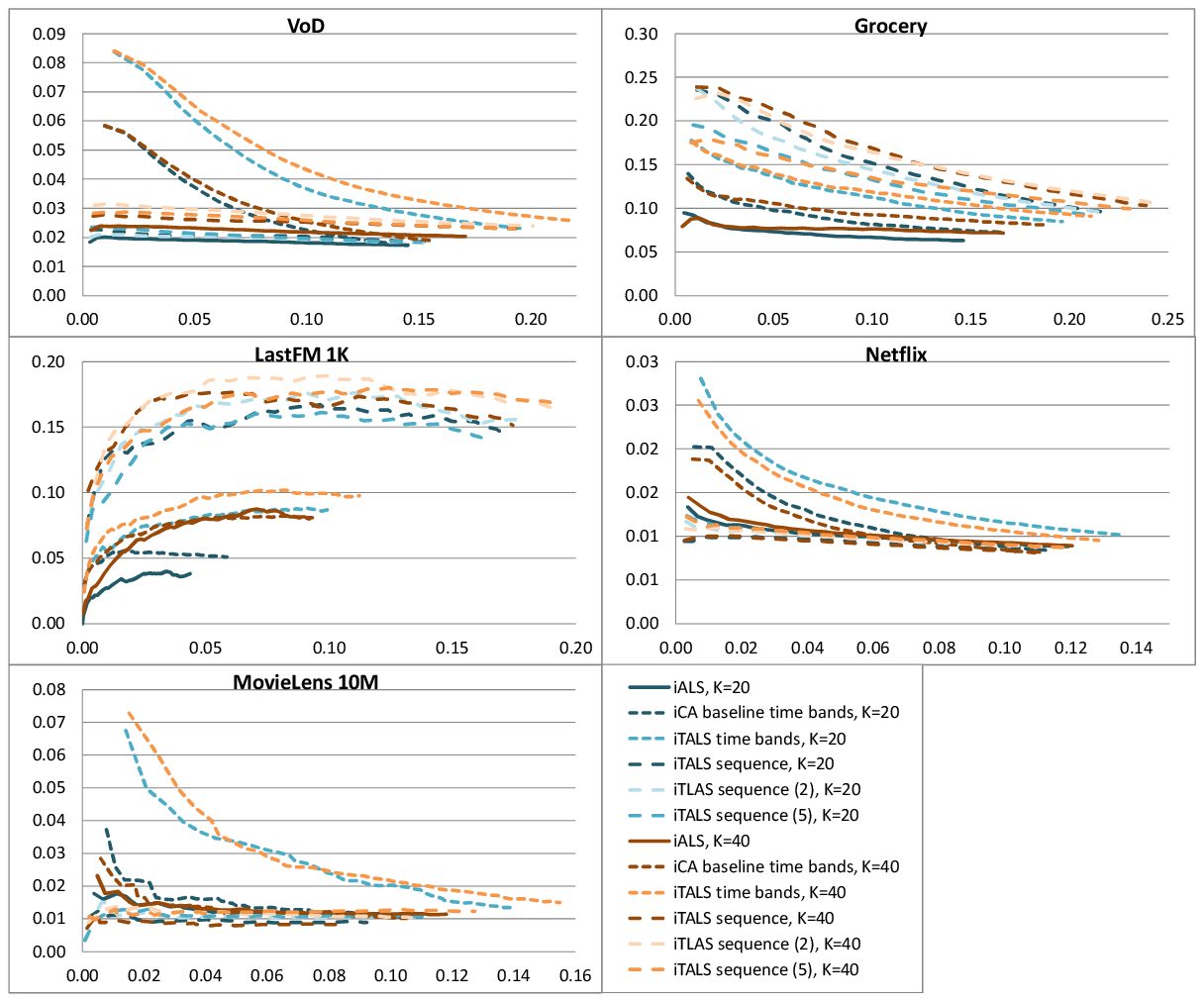}
\caption{Precision--recall curves for all datasets and algorithms using factorization with $K=20$ (blue) and $K=40$ (orange) factors. The $y$ axis corresponds to precision and $x$ to recall.}
\label{fig:curve}
\bigskip
\centering
\includegraphics[width=4.9in]{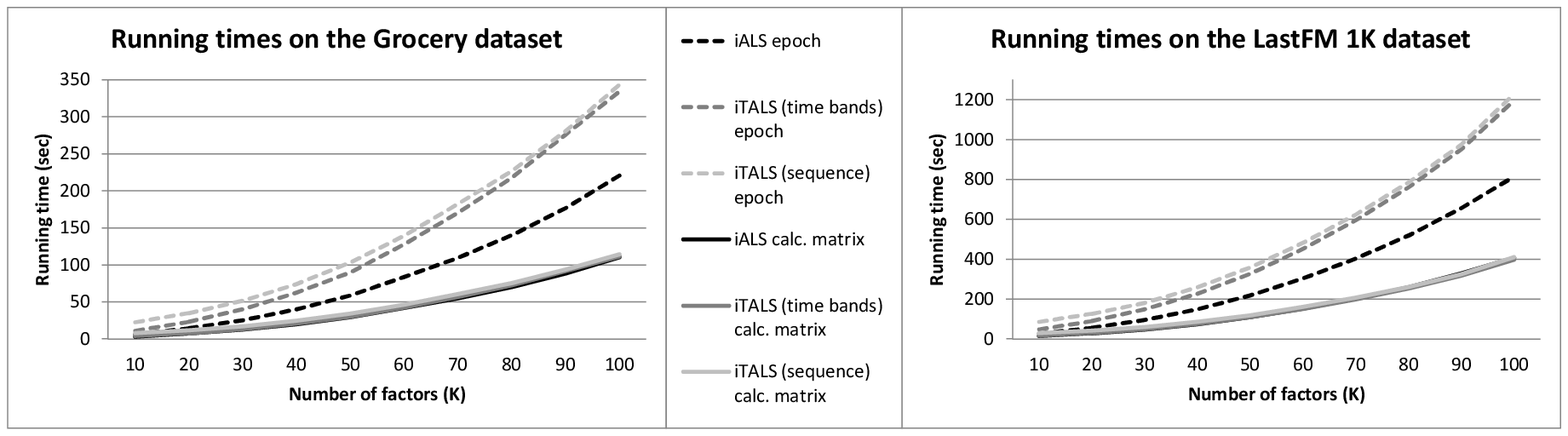}
\caption{Running times of iTALS compared to iALS on the Grocery and LastFM 1K datasets.}
\label{fig:runtime}
\end{figure}

Table~\ref{tab:recall} contains recall@20 values for every experiment. Recall@20 is important in practical application as the user usually sees maximum the top 20 items. Using context, the performance is increased overall. The selection of the appropriate context is crucial. In our experiments seasonality improved performance on all datasets. The sequentiality patterns caused large improvements on the Grocery and LastFM 1K datasets (significantly surpassed the results with the seasonality) but did not increased performance on the movie databases (VoD, Netflix, MovieLens). By including seasonality the performance is increased by an average of $30\%$ for the VoD data. This agrees with our assumption that the VoD consumption has a very strong daily repetitiveness and the behavior in different time bands can be well segmented. The results increased by an additional $35\%$ when we used iTALS instead of the context-aware baseline. The genre of the previously watched movies can also improve performance, however its extent is only around $10\%$. On the other two movie datasets iCA did not improve the performance significantly. We assume that this is due to the explicit--implicit transformation because the transformed implicit feedback is more reliable and also results a sparser tensor. The iTALS using seasonality however could achieve $30\%$ and $80\%$ improvement on Netflix and MovieLens respectively.

Inclusion of the sequentiality patterns increased the performance on Grocery and LastFM 1K datasets by more than $90\%$ and $300\%$ (compared to iALS, recall that no sequential iCA baseline is calculated). Interestingly, the model using the last category is the best with 20 features, but with 40 features the model using last two categories becomes better. We conjecture that this is connected to the greater expressive power of the model with more features. With seasonality the performance also improved by more than $50\%$ and $75\%$, respectively, on these datasets. We expected that the usage pattern learning will perform better on Grocery and LastFM 1K datasets than on the movie datasets as sequentiality is rather important in shopping and music listening than seasonality.

Figure~\ref{fig:curve} shows the precision--recall curves. The order of the performance of the algorithms is the same as with the recall@20. Observe that the distance between the curves of the iTALS variants and the curve of the iALS is larger when we use 40 features. Recall that the feature vectors of the context works as a reweighting of the user--item relation. If the resolution of this relation is finer, the reweighting can be more efficient and each factor describes a more specific item property, so the behavior in different context can be described more specifically. Thus, increasing the number of features results in larger performance increase for the context-aware iTALS variants than for iALS.

\subsection{Running times}\label{sec:runtimes}
We compared the running times of the iTALS and iALS algorithms in terms of $K$ (see Figure~\ref{fig:runtime}). The experiments were run on a laptop with an Intel Core i5 2410M 2.3GHz processor using only one core. We depict only curves for 2 datasets, since others are similar. We made several runs for each $K$; the median of the epoch running times are shown (dashed lines). The solid lines show the computation time for one feature matrix. Observe that iTALS scales quadratically with $K$ as iALS; the (re)computation time of one feature matrix is basically the same. Since iTALS recomputes more feature matrices its running time per epoch is larger. Importantly, even if the number of context-states is large (as with the sequential iTALS on LastFM 1K) the $O(K^2)$ part of the complexity remains dominant. This is because the number of non-zero elements in $T$ is much larger than the number of different items/users/contex-states in every case where the usage of context-aware approaches is justified. 
\section{Conclusion}\label{sec:conc}
In this paper we presented an efficient ALS-based tensor factorization method for the context-aware implicit feedback recommendation problem. Our method, coined iTALS, scales linearly with the number of \emph{non-zeroes} in the tensor, thus it works well on implicit data. We presented two specific examples for context-aware implicit scenario with iTALS. When using the seasonality as context, we efficiently segmented periodical user behavior in different time bands. When exploiting sequentiality in the data, the model was able to tell apart items having different repetitiveness in usage pattern. These variants of iTALS allow us to analyze user behavior by integrating arbitrary contextual information within the well-known factorization framework. Experiments performed on five large datasets show that proposed algorithms can greatly improve the performance. Compared to iALS and iCA, our algorithm attained an increase in recall@20 up to $300\%$ and $35\%$.

One should, however, avoid creating a high dimensional tensors because the number of non-zero elements remains the same no matter how many context types are integrated; so tensors with more dimensions become sparser and thus the results may be poorer than with only a few context dimensions used.
Our work opens up a new path for context-aware recommendations in the most common implicit feedback task when only the user history but no rating is available. Future work will include the characterization of the relation between reweighting, context features and the number of features ($K$) as well as the design of further context-aware iTALS-based recommendation algorithms. 

\section*{Acknowledgment}

Authors thank G\'abor Tak\'acs for his very useful comments on the paper.

\small
\bibliographystyle{splncs}
\bibliography{citations}

\end{document}